\documentclass[runningheads]{llncs}
\usepackage[T1]{fontenc}
\usepackage{graphicx}
\begin{document}
\title{Domain adaptation for handwriting trajectory reconstruction from IMU sensors}
\titlerunning{DA for handwriting IMU trajectory reconstruction}
%
\author{Florent Imbert\inst{1} \and
Romain Tavenard\inst{2} \and
Yann Soullard\inst{2} \and
Eric Anquetil\inst{1}}

\authorrunning{F. Imbert et al.}
%
\institute{IRISA, Universite de Rennes, INSA Rennes, Rennes, France   \and
IRISA, Universite Rennes 2, Rennes, France \\
\email{florent.imbert@irisa.fr}\\
}
\maketitle              
\begin{abstract}
Digital pens are commonly used to write on digital devices, providing the handwriting trace and enhancing human-computer interation. This study focuses on a digital pen equipped with kinematic sensors, allowing users to write on any surface while simultaneously preserving a digital trajectory of handwriting. This technology holds significant potential as a valuable educational tool, particularly in classrooms where it can facilitate the process of learning to write.
A major issue is based on the difference in captured signals between adults and children. For similar handwriting trace, we have large differences in sensor signals due to differences in speed and confidence in the handwriting gesture of children. 
To address this, we investigate a domain adaptation approach to build a unified intermediate feature representation aimed at facilitating the trajectory reconstruction. 
We demonstrate the interest of domain adaptation methods in leveraging existing knowledge for application in different contexts. Specifically, we compare our domain adaptation approach with two other methods: training the model from scratch and fine-tuning the model.

\end{abstract}
\section{Introduction}

This work focuses on reconstructing digital handwriting trajectories using the Digipen stylus \cite{KIHT}. The Digipen is equipped with 2 accelerometers, 1 gyroscope and a force sensor, and it can be used to write on any surface. While Inertial Measurement Units (IMU) are commonly used in tracking systems due to their cost-effectiveness, their signals are often imprecise due to high noise levels, presenting challenges. This issue is particularly prominent in the context of handwriting trajectory reconstruction, where precision is crucial, especially for e-learning purposes that require accurate feedback.

Although the Digipen stylus can be used on any surface and by different users, state-of-the art works \cite{Swaileh2023,wehbi2022surface} have experimented handwriting reconstruction approaches from data written by adults on tablets. Using the Digipen in another experimental context, e.g. on data written by children, or on another surface, e.g. on paper, leads to very different input signals. On the one hand, children’s handwriting are of variable speed depending of the assertiveness in the handwriting gesture. On the other hand, handwriting on paper produces noisier signals due to friction than when the user writes on a tablet. This leads us to consider a domain adaptation method for dealing with the different domains of data.

In this context, we present a domain adaptation approach designed to enhance the adaptability of our model to deal with the different data sources. Initially trained on data acquired from tablets by adults, our model aims to effectively handle a additional types of data: handwriting acquired from tablets by children. 
In fact, while collecting labelled adult data on a tablet is not a problem, it's more complicated to collect data from children (contact a school).
This approach is expected to broaden the applicability of handwriting trajectory reconstruction, making it more versatile and robust across different user groups.
To our knowledge, no domain adaptation method addresses handwriting trajectory reconstruction from different sources, e.g.  from adults vs children.

\section{Related works}

\subsubsection{Handwriting Trajectory Reconstruction} 
Traditional approaches \cite{Miyagawa,7574685}, which do not rely on deep learning, often struggle with sensor noise and error accumulation. These methods typically involve a series of preprocessing steps, such as applying low-pass filters to remove high-frequency noise and using coordinate transformation matrices to adjust for sensor orientation. However, the cumulative effect of noise and small errors during these preprocessing steps can lead to significant inaccuracies in the reconstructed trajectories over time.

Deep learning methods offer a more advanced approach to handling IMU data by learning and adapting to noise patterns, potentially providing more accurate trajectory reconstructions. For instance, \cite{ott2022joint} applied multi-task learning for joint classification and trajectory reconstruction. \cite{MagHacker} explored magnetic signals, while \cite{wehbi2022surface} addressed the Stabilo Digipen, using linear interpolation to align pen and tablet signals as preprocessing in training. This alignment is crucial for matching sequences of variable sizes and having a point-to-point matching for training the network. A Convolutional Neural Network is used for online handwriting trajectory reconstruction.

\begin{figure}[ht]
\begin{center}
\includegraphics[width=\textwidth]{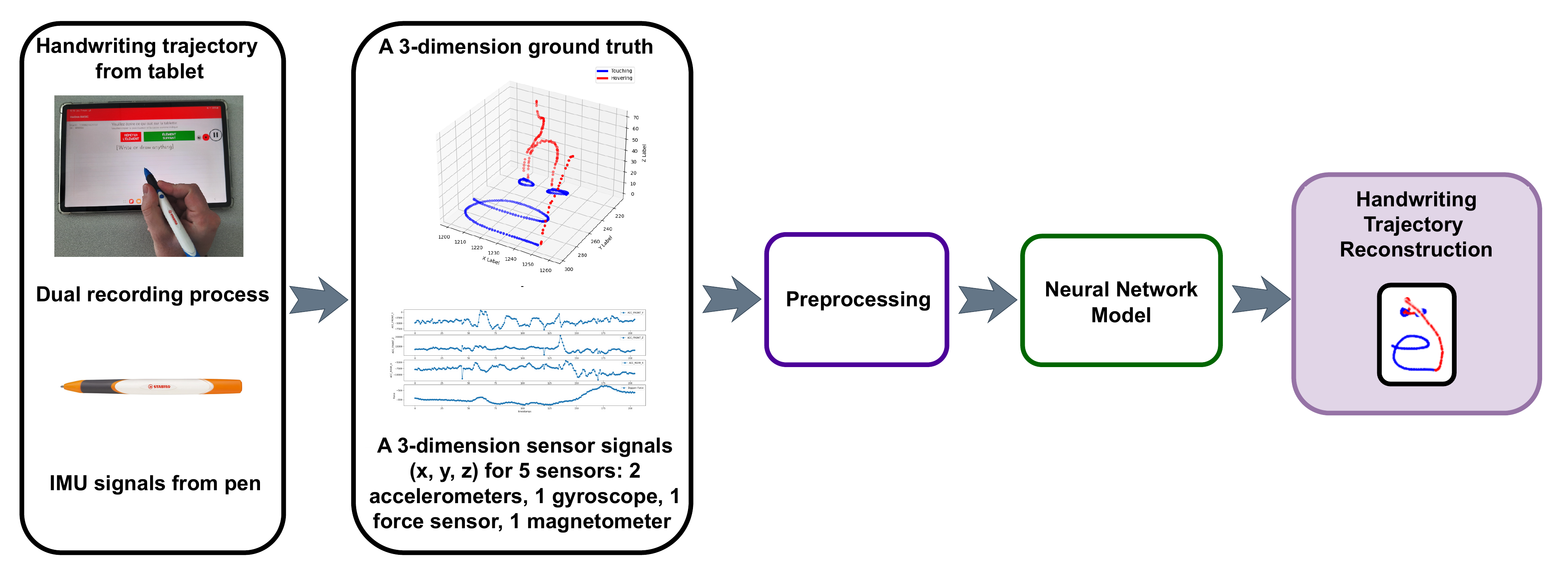}
\end{center}
\caption{Pipeline of handwriting reconstruction.}
\label{fig:pipline}
\end{figure}

Recently, \cite{Swaileh2023} improved on \cite{wehbi2022surface} by proposing a complete pipeline for online handwriting trajectory reconstruction (Fig \ref{fig:pipline}). Specifically, they used a specific preprocessing based on the Dynamic Time Warping (DTW) to align the ground truth with the input signal, which has the advantage of preserving the dynamics of handwriting, unlike the linear interpolation used in \cite{wehbi2022surface} and use a TCN model to extract local context information.

\subsubsection{Domain adaptation methods}
Domain adaptation (DA) adjusts a model trained on one domain (source) to perform well on a different but related domain (target), mainly when labeled data are rare or unavailable in the target domain. The goal is to minimize domain shift, or differences in data distribution, between the source and target domains. DA can be:
\begin{itemize}
    \item Supervised: Labeled data from the target domain are available.
    \item Semi-Supervised: Both labeled and unlabeled data from the target domain are available.
    \item Unsupervised: Only unlabeled data from the target domain are available.
\end{itemize}
Domain adaptation approaches can be broadly categorized into divergence-based, adversarial-based, and reconstruction-based methods. 
Divergence-based DA minimizes statistical differences between source and target distributions using measures like Maximum Mean Discrepancy (MMD) or Kullback-Leibler (KL) divergence \cite{shen2018wasserstein}. These methods focus on reducing the distribution gap to improve model performance on the target domain. Adversarial-based DA, inspired by Generative Adversarial Networks (GAN), employs a discriminator and feature extractor to create domain-indistinguishable features through adversarial training \cite{dann}. This approach leverages the power of adversarial learning to align features from different domains, making them indistinguishable to a domain classifier. 
Reconstruction-based DA, uses autoencoders to maintain reconstruction ability across domains, thereby creating robust, domain-agnostic feature representations \cite{ghifary2016deep}. By ensuring that features can be accurately reconstructed regardless of the domain, these methods enhance the model's robustness and adaptability. Together, these approaches help models generalize better to target domains by aligning data distributions, creating domain-invariant features, or ensuring robust feature representations, thereby improving performance and reliability in diverse settings.

\section{DANN-based method for handwriting reconstruction}
To get a shared feature space for different data sources, we explore a an adversarial-based DA method, named Domain-Adversarial Training of Neural Networks (DANN).  

\subsubsection{Motivation}
\begin{figure}[ht]
\centering
\includegraphics[width=0.9\textwidth]{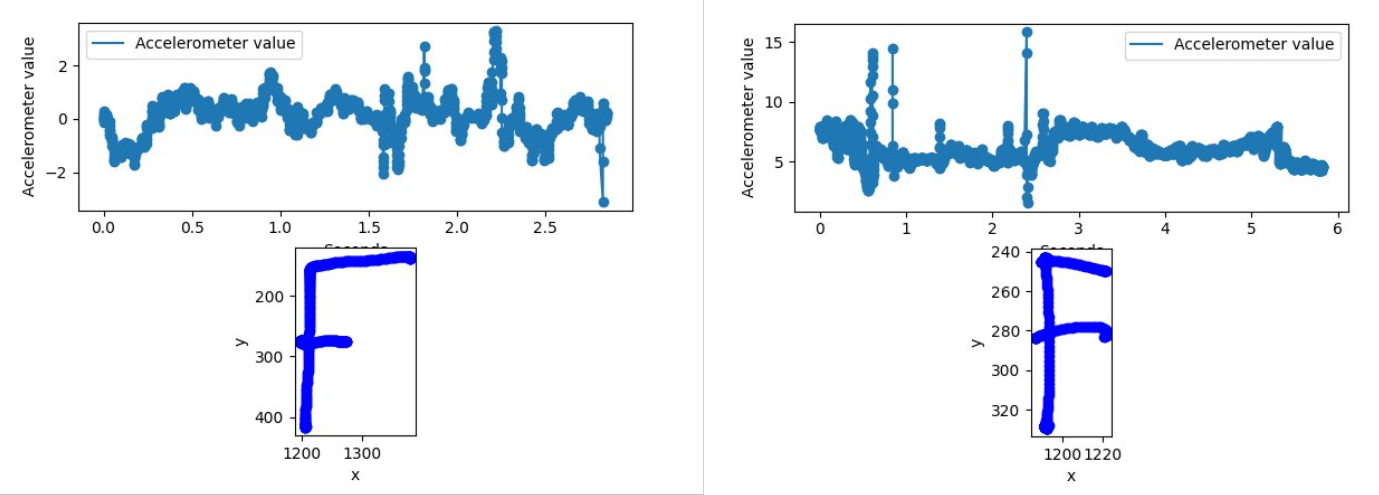}
\caption{Visualization of the x component of the Digipen's rear accelerometer over time in seconds, from left to the right: a F from adult on tablet, a F from children on tablet.We notice that the same pattern is written but not with the same intensity due to the different level of automation of handwriting (adult vs child)}
\includegraphics[width=0.75\textwidth]{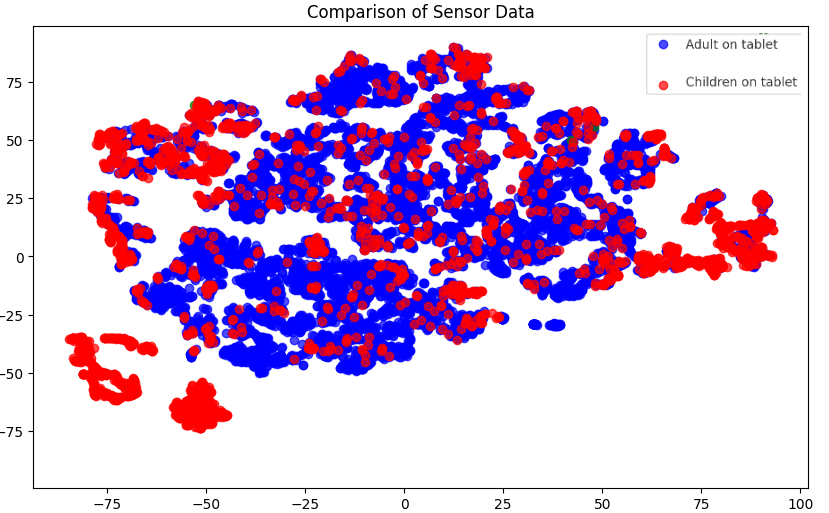}
\caption{Data visualization with the Multidimensional Scaling (MDS) method, we can see that children's data on tablets (in red) takes on a wider range of values due to their handwriting which is still being learned.}
\label{fig:data_visu}
\end{figure}

The variability of handwriting sources complicates the task of handwriting reconstruction. Children handwriting poses a unique challenge due to the ongoing development of graphomotor skills, resulting in dynamic and inconsistent handwriting patterns as children learn to write (Fig. 2). This translates into longer signal sequences than adults and a wider range of possible values (Fig. \ref{fig:data_visu}). 
This difference makes it difficult the handwriting reconstruction using a model trained on tablet-acquired data.

\subsubsection{DANN as a solution }
Domain-Adversarial Training of Neural Networks \cite{dann} are specialized neural network architectures designed to address the challenge of domain shift, where a model trained on one domain (source) is expected to perform well on a different but related domain (target). These networks operate by learning features that are domain-invariant, meaning they are useful and generalizable across both domains. This is typically achieved through a shared feature extractor on which additional components are built: a domain classifier and a task-specific classifier. The domain classifier is trained to determine the domain of the input data, whereas the task-specific classifier focuses on predicting the label of the dedicated task. In our context, the task-specific classifier is trained to reconstruct the handwriting trajectory. 

Training involves a twist (Fig. \ref{fig:dann}): the domain classifier's gradients are reversed during backpropagation, which encourages the feature extractor to generate features that are indistinguishable between domains, thus fooling the domain classifier. This technique, known as adversarial training, helps the network to minimize the representation gap between the source and target domains, leading to better performance on the target domain without requiring extensive labeled data from it.

\subsubsection{Application}
\begin{figure}[ht]
        \centering
        \includegraphics[width=\textwidth]{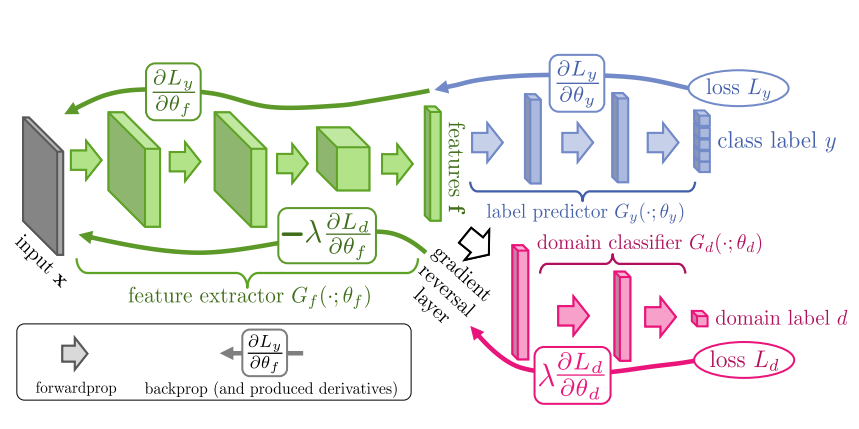}
        \caption{DANN from \cite{dann}.}
        \label{fig:dann}
\end{figure}

\begin{figure}[ht]
        \centering
        \includegraphics[width=0.7\textwidth]{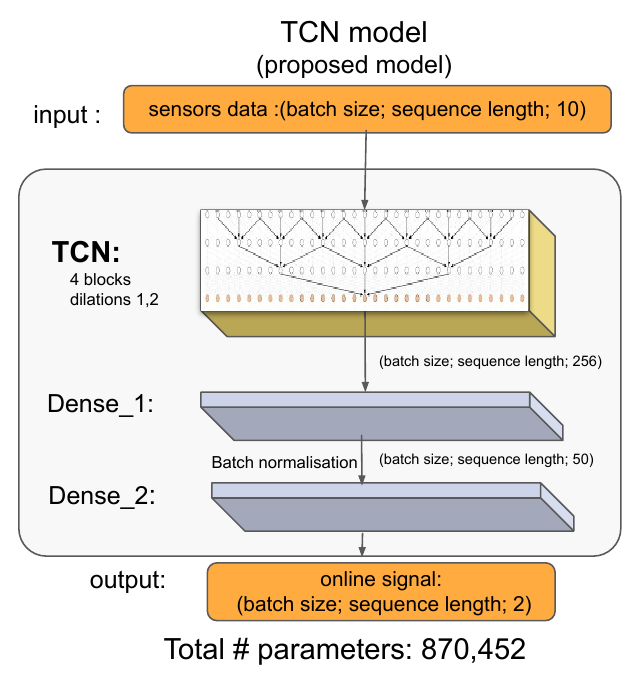}
        \caption{TCN model from \cite{Swaileh2023}. Also named baseline model (BM).}
        \label{fig:tcn}
\end{figure}

In this work, the reconstruction part of the DANN is the TCN-based network from \cite{Swaileh2023} which is currently the state-of-the art model for handwriting reconstruction from the Digipen sensors (Fig. \ref{fig:tcn}). We named it baseline model (BM) in the following. Then we have slice the baseline model as follows: the 4 blocks of the non-causal TCN is the feature extractor (in green in Fig. \ref{fig:dann}). Each TCN block is composed of 2 convolutions with dilation 1 and 2 respectively and a kernel size of 3. The next two dense layers refer to the label predictor (in blue in Fig. \ref{fig:dann}) which in our context corresponds to the trajectory reconstruction.
The domain classifier (in pink in the Figure \ref{fig:dann}), is made up of a max polling layer followed by two dense layers.

In our case study, we aim to transition from adult handwriting on a tablet (source) to children handwriting on a tablet (target). During training, we provide mixed matches to two model branches: (1) the feature extractor and the label predictor (green + blue in Fig. \ref{fig:dann})  pre-trained on adult data, and (2) a feature extractor and domain classifier (green + pink in Fig. \ref{fig:dann}).

\section{Experimental results}
\begin{figure}[ht]
\centering
\includegraphics[width=0.8\textwidth]{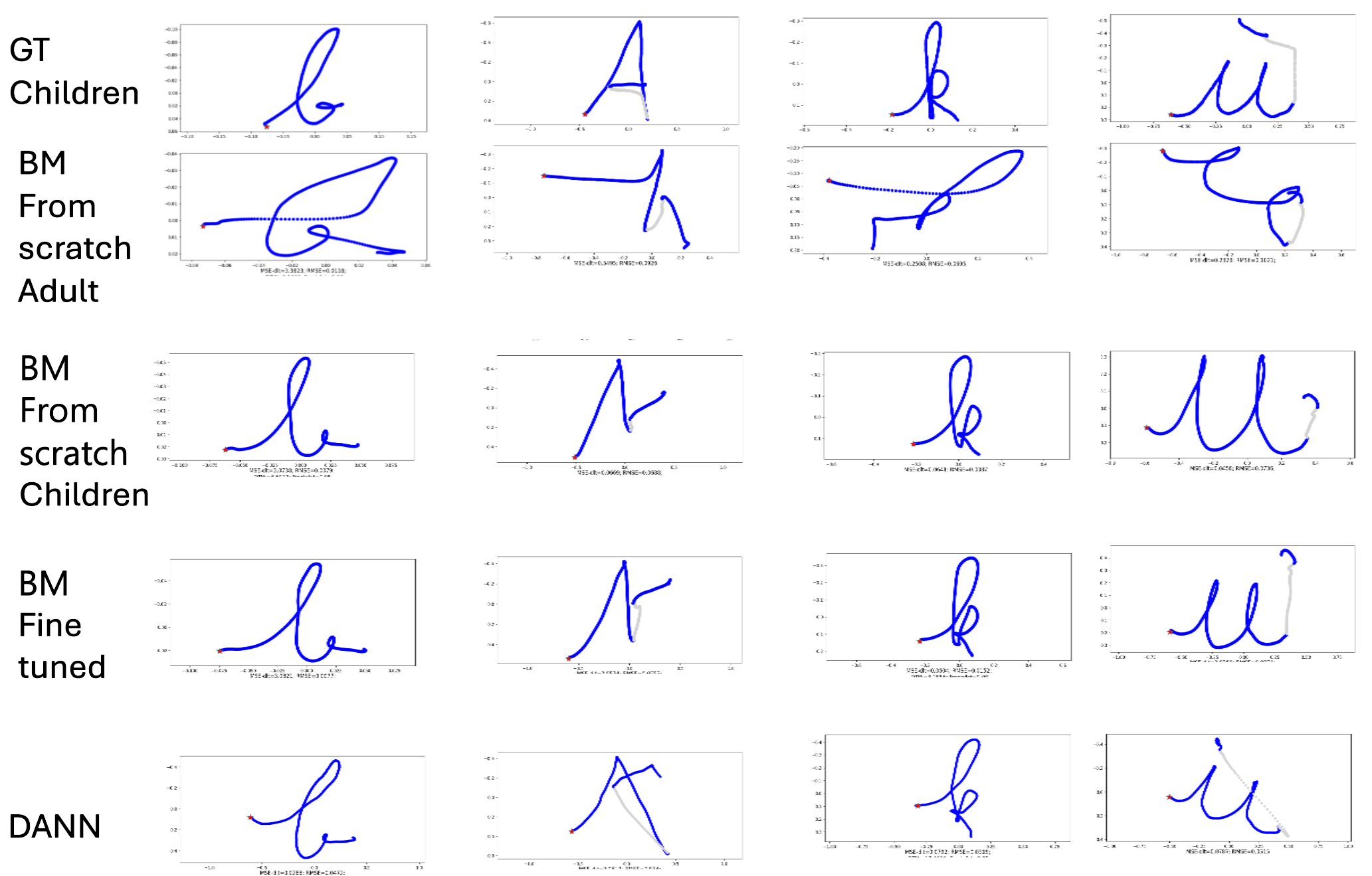}
\caption{Comparison of reconstruction methods on children test data, on the first line the ground truth, on the second the Baseline Model trained on adult data, on the third the Baseline Model trained on child data, then the Baseline Model trained on adult data and fine-tuned on child data, on the last line the DANN pretrain on adult data.}
\label{fig:Results}
\end{figure}

We experiment our approach on two datasets, one for adult tablet data (9629 samples), another for children tablet data (3910 samples). Each datasets contains characters, words, word groups, equations, and shapes. Following \cite{Swaileh2023}, we compute the Fréchet distance to evaluate the quality of reconstruction on children's characters.
We trained a DANN with children and adult data, and compared it qualitatively (Fig. \ref{fig:Results}) and quantitatively (Table \ref{tab:evaluation_metrics}) to the following methods: baseline model trained on adult data and fine-tuned on children data and the baseline model trained from scratch on children data.

\begin{table}[ht]
    \centering
    \caption{Comparison of reconstruction methods between training from scratch, fine tuning and domain adaptation method on children test data with the Fréchet distance.}
    \label{tab:evaluation_metrics}
    \begin{tabular}{|l|c|c|c|c|}
        \hline
        Model & \multicolumn{3}{|c|}{BM} & DANN \\
        \hline
        Pretraining Data &   &   & Adult & Adult \\
        \hline
        Training Data & Adult & Children & Children  & Children \\
        \hline
        Fréchet distance  & 0.470 & \textbf{0.345} &  0.348 & 0.349 \\
        \hline
    \end{tabular}
\end{table}

\begin{table}[ht]
    \centering
    \caption{Comparison of reconstruction methods between training from scratch, fine tuning and domain adaptation method on adult test data with the Fréchet distance.}
    \label{tab:evaluation_metrics_adlut}
    \begin{tabular}{|l|c|c|c|c|}
        \hline
        Model & \multicolumn{3}{|c|}{BM} & DANN \\
        \hline
        Pretraining Data &   &   & Adult & Adult \\
        \hline
        Training Data & Adult & Children & Children  & Children \\
        \hline
        Fréchet distance & \textbf{0.332} & 0.378 &  0.386 & \underline{0.364} \\
        \hline
    \end{tabular}
\end{table}

Table \ref{tab:evaluation_metrics} shows  that methods integrating children data perform the best overall. Specifically, all the methods trained on children's data improve the Baseline Model trained only on adult data. The advantage of DANN is that it keeps a common representation of adult and children's features, and table \ref{tab:evaluation_metrics_adlut} shows that it performs well in both areas, making it a 2-in-1 solution. Unlike from scratch and fine tuning models (Table \ref{tab:evaluation_metrics}), which lose out on performance in the domain where they are not learned (Table \ref{tab:evaluation_metrics_adlut}). Regarding the qualitative analysis (Fig. \ref{fig:Results}), we observe that the trajectory reconstruction using the DANN is quite satisfactory and it seems closest to the ground truth than using the other approaches on those examples. The unique representation shared by the two data sources seems help the model the model in the trajectory reconstruction, especially on the hovering part.
Additionally, this work is still in progress and the DANN has potential for further improvement. 

\section{Conclusion}

This paper shows the benefits of retaining knowledge from one domain (adults on tablet) and moving on to a second (children on tablet).
Our areas for improvement include optimizing the lambda parameter to adaptively manage the weight of each branch over time, which could enhance DANN performance. Another area for improvement involves creating DANN batches and studying factors such as padding.
We will also investigate domain adaptation from data acquired on tablet to data written on paper.

\begin{credits}
\subsubsection{\ackname} This project is financed by the KIHT French-German bilateral ANR-21-FAI2-0007-01 project and these four partners, IRISA, KIT, Learn \& Go and Stabilo. This work was performed using HPC resources from GENCI-IDRIS (Grant 2021-AD011013148)

\subsubsection{\discintname}
The authors have no competing interests to declare that are
relevant to the content of this article. 
\end{credits}

%
%
%
\bibliographystyle{splncs04}
\bibliography{bib}

\end{document}